# An adversarial attack approach for eXplainable AI evaluation on deepfake detection models

Balachandar Gowrisankar, Vrizlynn L.L. Thing
*ST Engineering, Singapore*
balachandar.gowrisankar@stengg.com, vriz@ieee.org

*Abstract*—With the rising concern on model interpretability, the application of eXplainable AI (XAI) tools on deepfake detection models has been a topic of interest recently. In image classification tasks, XAI tools highlight pixels influencing the decision given by a model. This helps in troubleshooting the model and determining areas that may require further tuning of parameters. With a wide range of tools available in the market, choosing the right tool for a model becomes necessary as each one may highlight different sets of pixels for a given image. There is a need to evaluate different tools and decide the best performing ones among them. Generic XAI evaluation methods like insertion or removal of salient pixels/segments are applicable for general image classification tasks but may produce less meaningful results when applied on deepfake detection models due to their functionality. In this paper, we perform experiments to show that generic removal/insertion XAI evaluation methods are not suitable for deepfake detection models. We also propose and implement an XAI evaluation approach specifically suited for deepfake detection models.

## 1. Introduction

Advancements in technology and computational resources have proliferated the creation and circulation of deepfakes on the Internet. Deepfake is a term used to denote fake images/videos/audios created using deep learning techniques. Tools used to create deepfakes have evolved over the years and as a result, humans are losing the capability to distinguish between real and fake content using naked eye. Thus, ML researchers are investing a lot of time and effort in this field to understand the process behind the creation of deepfakes and develop models to detect them. Cross Efficient Vision Transformer [1], XceptionNet [2] and BA-TFD [3] are few benchmark deep learning models proposed in the recent past to combat the spread of fake content on the web. While researchers continue to develop complex detection models, humans are finding it difficult to blindly believe a model's decision. Model evaluation metrics like accuracy, precision, recall etc. are not proving enough to gain a person's trust on the model. This lead to the development of XAI. These are tools designed to help a human understand the reasons behind a model's decision. LRP [4], Grad-CAM [5], LIME [6] and SHAP [7] are some popular XAI tools used in practice.

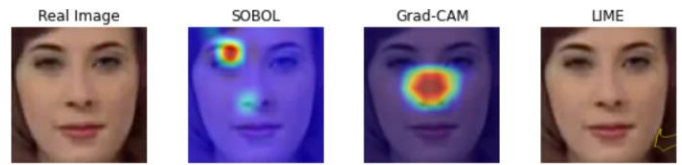

Fig. 1: Ambiguities among XAI tools on XceptionNet for a real image

With a plethora of tools available to the public, it is important to select the right tool for a model. Naturally, we may think that all tools will provide similar results and aspects like complexity, speed etc. should be the ones setting them apart. An example of applying SOBOL, Grad-CAM and LIME on XceptionNet for a real image is shown in Figure 1. In the case of SOBOL and Grad-CAM, the pixels marked in red are the ones which contribute highly to a the model's decision while LIME denotes the responsible pixels with a yellow boundary. In SOBOL's explanation, the red regions are aligned more towards the left eye whereas they are pointed towards the nose in Grad-CAM. LIME highlights a part of the hair region in its explanation (shown here at the bottom right corner). This clearly shows that ambiguities exist among XAI tools and all of them may not highlight the same set of pixels for a given image and model. Thus, there is a need to evaluate the results of these tools to determine the right tool for a model.

Generic XAI evaluation methods like [4], [8], [9], [10] and [11] focus on aspects like pixel/segment removal or insertion to assert the correctness of salient pixels. These methods can be useful for tasks like object detection where the pixels contributing to a model's decision can be removed/inserted to check its effect on the model's prediction. Figure 2 shows how removal of salient pixels can help in verifying if the model's prediction for 'goldfish' class drops or not [9]. Evaluating XAI tools by removing/inserting salient pixels provide meaningful results for object detection tasks since the model learns object specific features during its training phase and thus removing the object itself from the image should intuitively decrease the model's prediction probability for that object. However, this is not the case with deepfake detectors. Recently, [12] and [13] demonstrated that deepfake detectors learn the patterns of visual concepts of a face. These visual concepts correspond to regions like the eyes, nose, mouth etc. Removing salient pixels in deepfake images will remove a visual concept from the person's face and thus, it may not be wise to judge a model's prediction

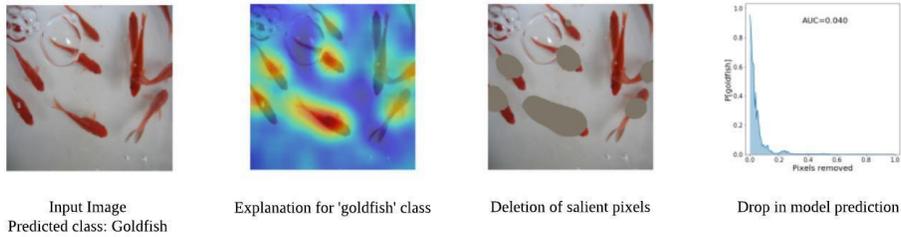

Fig. 2: Evaluation of salient pixels for object detection task

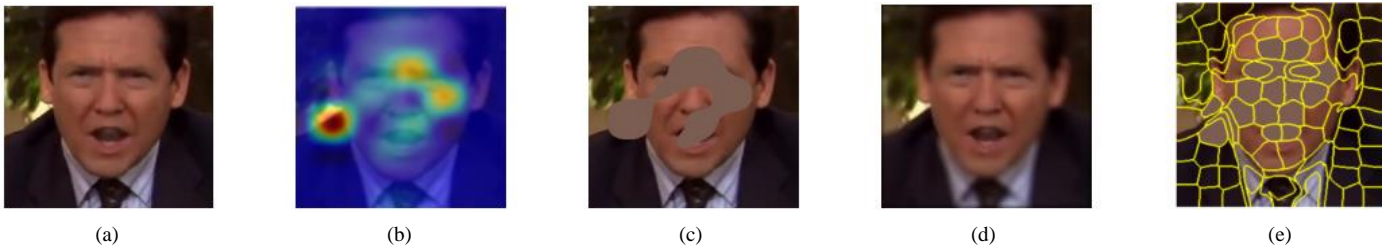

Fig. 3: Illustration of changes done to images for various evaluation methods (a) Fake image (b) Explanation of fake image (c) Deletion: Top 20% pixels replaced with per-channel mean (d) IAUC: Top 20% pixels inserted on a blurred image (e) IROF: Top 20 segments replaced with per-channel mean

without feeding in the full face. Besides, the model's prediction probability may not even decrease as expected on removing salient pixels. We perform experiments with different removal/insertion XAI evaluation methods on deepfake detectors to demonstrate these limitations.

In the context of deepfake images, a fake image indicates an image of a person wherein a portion of or the entire face has been tampered. The tampering may be done manually using popular photoshop applications or artificially generated using AI models. A real image is an image of a person whose face has not been tampered. While classifying faces, an image that has a prediction probability close to 0 is classified as a real image while an image that has a prediction probability close to 1 is classified as a fake image.

An intuitive way to evaluate the faithfulness of tools on deepfake detectors is to check if the salient visual concept of a real image can be used to flip the prediction of a fake image Ideally, we must be able to misclassify fake images after knowing the visual concepts used by a model to predict real images. Using this simple intuition, we propose an adversarial attack evaluation approach based on visual concepts. For a given real-fake image pair, we first identify the visual concept highlighted by a tool on a real image and perturb the same concept in the corresponding fake image to generate an adversarial fake image. Since different tools are bound to highlight different visual concepts, we can rank the tools by comparing the ability of the adversarial fake images to reduce the model accuracy by a significant amount. Such an approach is not possible for other image-classification related tasks since the salient pixels of images belonging to different classes cannot be correlated.

To summarize, the main contributions of this work are listed as follows.

- We performed experiments to demonstrate the inefficiencies of generic removal/insertion XAI evaluation methods on deepfake detection models.

- We proposed and implemented an XAI evaluation method specifically suited for deepfake detection models.

The rest of the paper is organized as follows. Section 2 reviews related works on the evaluation of XAI tools. Section 3 highlights the details about various XAI tools and the adversarial attack method used in this paper. The design of our proposed evaluation approach is discussed in Section 4. The experimental setup and results are presented in Section 5 and Section 6. Section 7 concludes the paper.

2. RELATED WORKS

This section highlights the limitations of current XAI evaluation methods for image classification models and presents the challenges of applying them to deepfake images. Since our work is closely aligned with attacking the salient regions of an input, this section also gives an overview of saliency-based adversarial attacks

2.1. XAI evaluation methods

Various methods have been proposed in the literature to evaluate XAI tools on models classifying images/videos. Pixel-flipping [4] was proposed by the authors of Layer-WiseRelevance Propagation tool (LRP). The images used for the experiments belonged to the MNIST handwritten digits dataset which were mostly black and white. Hence, the method involves a state flip of the binary pixel values sorted in descending order based on the explanation and observing the rate at which the model's prediction changes. However, a state flip of pixel values would not be applicable in scenarios where the images contain colors other than black and white.

Deletion [8] was a generalization of the pixel-flipping approach. Starting with the original image, the 'k' most important pixels are removed one by one from the original image based on the explanation and the model's prediction is computed. The intuition here is that the model's prediction should drop since the important features are removed. The removed pixels are replaced with a constant value (per-channel mean, black etc.), or picked from a uniform random distribution or blurred with an average value based on the neighbouring pixels. Insertion Area Under Curve (IAUC) [9] is another metric similar to Deletion. Instead of removing pixels, the authors start with a blurred version of the original image and insert 'k' most important pixels based on the model's explanation. Unlike Deletion, the model's prediction is expected to rise after the important pixels are inserted.

Iterative Removal Of Features (IROF) [10] divides the image into a number of segments using slic algorithm of skimage. The segments are then ranked based on the explanation and the top 'k' segments are deleted from the image. The deleted segments are replaced using any of the methods described for Deletion. The computation of IROF takes lesser time when compared to Deletion and IAUC.

Figure 3 illustrates the changes done to images for the evaluation methods discussed so far. Notice that Figure 3(c), 3(d) and 3(e) produce images that are visually dissimilar from the original one. In addition to this, all these methods suffer from Out-Of-Distribution (OOD) issue. OOD arises when a model is shown testing data that does not belong to the training data distribution. The prediction given by the model on such data should not be trusted upon. By inserting pixels using different replacement methods, the resulting data may not lie within the boundary of the original training distribution. As a result, it cannot be ascertained whether the drop/rise in the prediction of the new image is due to removal of important features or because of the shift in distribution. Gomez et al. [14] proved that distortions to an image like the one shown in Figure 3(d) does introduce OOD issues and leads to unexpected behaviour of the model. Further, there may exist a local correlation among pixels even after removal/insertion of important features which will still allow the model to guess the prediction correctly. In the case of Figure 3(c), if the actual value of the removed pixel is very close to the average value, then not much of information is removed from an image. While segmentation removes the problem of local correlation in IROF, the issue of OOD still persists. In our work, only a small magnitude of noise is added to an image to ensure that it remains visually imperceptible from the original one. We also make use of segmentation in our approach to prevent local correlation among pixels. In other words, noise will be added to all pixels that represent similar features.

Pointing game [15] measures how well an explanation tool can highlight pixels in a pre-annotated area such as a bounding box. The pointing game score is defined as the ratio of total number of hits to the total number of all samples. It is useful in a domain where the ground truth for the salient area is known beforehand (for example, object detection tasks). This method cannot be applied on deepfakes. This is because we as humans cannot ascertain which areas contribute to the real/fake class and thus, cannot pre-annotate any area in a real/fake image to check if the salient pixels fall within that area.

RemOve And Retrain (ROAR) [11] is an extension of Deletion and overcomes the OOD limitation. The authors of ROAR argue that it is necessary to retrain the model after deletion to judge whether the reduction in model performance is due to the distribution shift or because of removal of important features. Their approach is to retrain the model using the images created after deletion and compare the accuracy between the original and retrained model. Ideally, a good tool should result in the retrained model having a lower accuracy than the original one. However, there are a few issues when it comes to adopting ROAR for practical purposes. Retraining consumes a lot of time and may demand more computational resources. The author of [10] notes that it is not feasible for research groups to use ROAR as an evaluation of eight tools based on ResNet50 may take around 241 days even with eight GPUs. Unlike ROAR, our work does not involve model retraining and hence is computationally inexpensive.

Max-Sensitivity [16] measures the degree to which an explanation changes on introducing random perturbations to the data. A small amount of noise is added to the image and the attribution map is recomputed on the perturbed image. The difference between the attribution maps of the original and perturbed image is measured. In the ideal case, an explanation should have less sensitivity. Having a higher sensitivity could possibly mean that the explanation is susceptible to adversarial attacks. While this metric may be useful in measuring the robustness of an explanation, it cannot be used to validate the faithfulness of an explanation.

Infidelity [16] compares the difference in model output after an arbitrary perturbation with the dot product of the perturbation vector and attribution map. The perturbation vector can be of two types: Noisy Baseline and Square Removal. The former deals with using a Gaussian random vector as the perturbation vector. Square Removal captures spatial information in the images by removing square patches from an image of predefined size. Once again, the perturbations used in this case are not constrained by any limit, thus making them perceptible.

Impact Coverage [17] adds an adversarial patch to the image such that it is misclassified. The explanation for the adversarial image is then computed. Intuitively, a good tool must highlight the adversarial patch in its explanation since it led to the misclassification. However, the adversarial patches used in this metric are image-independent and visually perceptible. Similar to Max-Sensitivity, this metric also requires the explanation method to be available at the time of evaluation. Our work focuses on creating adversarial images that are visually imperceptible from the original ones and does not require re-computation of the explanation on the adversarial images.

*2.2. Saliency-based adversarial attacks*

Works that confine the adversarial attack perturbation to the salient regions of an input have also been published in recent literature. However, unlike our work, their primary motivation was not to evaluate the salient regions. Dong et al. [18] showed that a superpixel based adversarial attack strategy guided by the results of CAM [19] was robust to image processing based defense and steganalysis based detection. Xiang at al. [20] formulated a local black-box adversarial attack technique to improve query efficiency and the transferability of adversarial samples. Instead of attacking the original image directly, they first perform a white-box attack on a surrogate model where the noise is confined to the regions highlighted by GradCAM [5]. The resulting image is then used as a starting point for the actual black box attack. Dai et al. [21] introduced a salient-based black-box attack method with the main intention of creating imperceptible adversarial samples. All these works utilize the salient regions of the same image for which they intend to produce an adversarial image. However, our work differs in the aspect that we bring into play the correlation of salient regions of images belonging to the opposite class. In other words, we identify the salient regions of a real image and attack those regions in the corresponding fake image to generate an adversarial fake image. This correlation is possible in deepfake images due to the similar orientation of faces in a real-fake image pair.

3. BACKGROUND

This section gives an insight on the XAI tools and adversarial attack method used in this paper.

*3.1. XAI*

XAI tools are divided into two types: model-specific and model-agnostic. Model-specific tools consider the internal structure and working of a model while making their calculations. Such tools can only be applied to a specific class of models like, for example, Convolutional Neural Networks (CNN). Model-agnostic tools on the other hand, do not require any knowledge about the structure of the model. They work by randomly perturbing the input data features and observe the model's performance for different perturbations. The following tools were selected for our experiments based on their compatibility with the chosen deepfake detection models:

(a) SOBOL [22]

It is a model-agnostic tool which leverages the use of a mathematical concept called SOBOL indices to identify the contribution of input variables to the variance of the output. A set of real-valued masks are drawn from a Quasi-Monte Carlo (QMC) sequence and then applied to an input image through a perturbation function (e.g. blurring) to form perturbed inputs. These inputs are forwarded to the model to obtain prediction scores. Using the masks and the associated prediction scores, an explanation is produced which characterizes the importance of each region by estimating the total order of SOBOL indices. One drawback of SOBOL is that it can be applied to image data only.

(b) eXplanation with Ranked Area Integrals (XRAI) [23]

XRAI combines Integrated Gradients (IG) [24] with additional steps to determine the regions of an image contributing the most to a decision. It performs pixel-level attribution for an input image using the IG method with a black baseline and a white baseline. It then over-segments the image to create a patchwork of small regions. XRAI aggregates the pixel-level attribution within each segment to determine its attribution density. Using these values, the segments are ranked and ordered from most to least positive. This determines the most salient regions of an image. Unlike SOBOL, XRAI is not model-agnostic since it makes use of gradients of a model.

(c) Random Input Sampling Explanations (RISE) [9]

Like SOBOL, RISE works by randomly perturbing the input and observes the changes in the model's predictions. Image perturbation is done by generating binary masks and occluding the corresponding regions of the image. The prediction scores of the perturbed images are used as weights to weight the importance of the mask. The intuition here is that masks that contain important pixels will be weighted more when compared to other masks. The weighted masks are then added up to produce a saliency map. The computation time for RISE is heavy since a lot of masks are required to explain a single image.

(d) Local Interpretable Model-agnostic Explanations (LIME) [6]

The authors of LIME proposed that it is not necessary to look at the entire decision boundary to explain a prediction but rather zooming into the local area of the individual data instance should be enough to provide a reasonable explanation. Numerous data samples are generated from the original input by randomly perturbing the features. These data samples are then weighted based on their distance from the original input. They are then fed to the model and their corresponding predictions are retrieved. The explanation provided by LIME lies around finding a simple linear model that fits the new data samples and their corresponding predictions.

(e) Gradient Class Activation Mapping (Grad-CAM) [5]

This tool is a slight modification of Class Activation Mapping (CAM) [19]. The CAM approach had a major drawback since the existing network had to be modified. Selvaraju et al. proposed a slight modification of CAM where the gradients of the classification score of the last convolutional layer can be used to identify salient parts of an image. The input image is run through the model and the last convolution layer's output and loss are retrieved. This is followed by calculating the gradient of that output with respect to the model loss. The average of the gradients is then multiplied with the last CNN layer's output to get the final saliency map. Grad-CAM is a model-specific tool as it can be applied to CNNs only.

*3.2. Adversarial attacks*

An adversarial attack on a model is the process of adding noise to the data such that the model reverses/changes its original prediction. Creating visually imperceptible adversarial samples is a challenging task as the amount of noise added cannot exceed a certain limit in order to maintain visual similarity. Adversarial samples can be created in two ways:

(a) White-box method

In this method, an adversary has complete access to the model parameters and gradients. Popular white-box methods to generate adversarial samples include Fast Gradient Sign Method (FGSM) [25] and Iterative-Fast Gradient Sign Method (I-FGSM) [26]. However, such type of attacks is not practical in real-world scenarios since an adversary may not have complete access to the model. The adversarial samples created using such type of attacks are not transferable, meaning that they can bypass only the model using which they were created.

(b) Black-box method

In this case, an adversary can only query the model for output and knows nothing else about the model. Since gradients cannot be directly retrieved from a model, gradient estimation techniques like Differential Evolution [27] and Natural Evolutionary Strategies (NES) [28] are a few black-box methods used to create adversarial samples. This attack is more challenging to execute when compared to its white-box counterpart due to the lack of information about the model. Since we believe that a good XAI tool should help to create better fake images without knowing a model's parameters, we will be dealing with this method in our evaluation approach. More details about its implementation is discussed in section 4.2.

## 4. METHODOLOGY

In this section, we present the details of our proposed evaluation framework. Our approach is to add noise to fake images in those visual concepts that contribute highly to the classification of real images as indicated by an explanation tool. For a given real-fake image pair, we first compute the explanation of the real image and identify the important visual concept. Next, we need to find a way to manipulate the same visual concept in the corresponding fake image. The orientation of faces in real-fake image pairs are similar. Hence, if we can derive the pixel indices of the salient visual concept of real images, those indices will provide us the regions for manipulation in the corresponding fake image. To implement this, we make use of image segmentation through slic algorithm of scikit-image package. The fake image is divided into a total of 100 segments. We then rank the segments based on the explanation of the real image to select the region for distortion. This is followed by the generation of adversarial sample using the selected segment. We present the implementation details of the ranking and adversarial image generation below. An illustration of our evaluation approach is shown in Figure 4.

### 4.1. Ranking segments

The fake image segments are ranked based on the mean importance approach followed by [10]. [10] showed that two segments can be compared by computing the mean importance of each segment according to a given explanation method. For a given image $i$, saliency map $E_i$ and a set of segments $\{S_i^l\}_{l=1}^{L}$ where $L$ is the total number of segments, the importance of a segment $S_i^l$ can be represented as:

$$Imp(S_i^l) = \frac{||E_i(S_i^l)||_1}{||S_i^l||_1}$$

In the above equation, $||x||_1$ represents the $L_1$ norm or the sum of the absolute values of a vector $x$. The segments are ranked in descending order based on their mean values and the indices $ind$ of the topmost ranked segment is retrieved. These indices correspond to the pixels to which noise will be added during adversarial image generation.

### 4.2. Adversarial sample generation

As discussed in section 3.2, we will be using a blackbox strategy to generate adversarial samples. Hussain et al. [29] used NES to generate adversarial deepfake videos. We adapt our implementation from their work. In order to maintain imperceptibility from the original image, it is necessary to ensure that the distortion between the original and adversarial image is minimal. The $L_p$ norm is a commonly used distance metric to measure the difference in noise between two images. Considering time-efficiency issues, Hussain et al. used the $L_\infty$ metric to create adversarial deepfake videos. Since our work

---

**Algorithm 1:** Adversarial image generation

**Input:** Fake image x, deepfake classifier F, indices ind
**Parameters:** Search variance $\sigma$, number of samples $n$, image dimensionality $N$, maximum iterations $itr$, maximum distortion $\epsilon$, learning rate $\alpha$
**Output:** Adversarial Image $x_{adv}$

for $i \leftarrow 1$ to $itr$ do
  if $F(x) = Real$ then
    $x_{adv} \leftarrow x$;
    return $x_{adv}$
  end
  $g \leftarrow 0$;
  for $j \leftarrow 1$ to $n$ do
    $u_j \leftarrow N(0_N, I_{N,N})$;
    $g \leftarrow g + F(x[ind] + \sigma u_j[ind])_{Real} \cdot u_j[ind]$;
    $g \leftarrow g - F(x[ind] - \sigma u_j[ind])_{Real} \cdot u_j[ind]$;
  end
  $g \leftarrow \frac{1}{2n\sigma}g$;
  $x[ind] \leftarrow x[ind] + \text{clip}_\epsilon(\alpha \cdot \text{sign}(g))$;
end
$x_{adv} \leftarrow x$;
return $x_{adv}$

---

closely follows their approach, we will be using the same metric for our implementation. In the $L_\infty$ metric, the maximum distortion of any pixel is constrained by a given threshold $\epsilon$. Thus, for a given fake image $x$, deepfake classifier $F$ and a set of indices $ind$, we mathematically define the problem of finding an adversarial fake image $x_{adv}$ as follows:

$$F(x_{adv}) = Real \text{ such that } ||x_{adv}[ind] - x[ind]|| < \epsilon$$

NES is a popular black-box technique used to estimate gradients. The main idea behind NES is to maximise the probability $F(x)_y$ for a given target class $y$. It achieves this by maximising the expected value of the function under a search distribution $\pi(\theta|x)$.

From [28], the gradient of expectation can be derived as given below:

$$\nabla_x E_{\pi(\theta|x)}[F(\theta)_y] = E_{\pi(\theta|x)}[F(\theta)_y \nabla_x log(\pi(\theta|x))]$$

Usually, a search distribution of random Gaussian noise is chosen around an image, that is, $\theta$ is set as $\theta + \sigma \delta$ where $\delta \sim N(0, I)$. However, instead of setting $n$ values $\delta \sim N(0, I)$, Hussain et al. used antithetic sampling to sample Gaussian noise for $i \in \{1, ..., \frac{n}{2}\}$ and set $\delta_j = \delta_{n-j+1}$ for $j \in \{\frac{n}{2} + 1, ..., n\}$ since this was shown to empirically improve the performance of NES. Estimating the gradient with a population of *n* samples yields the following variance reduced gradient estimate:

$$\nabla E[F(\theta)] \approx \frac{1}{\sigma n} \sum_{i=1}^{n} \delta_i F(\theta + \sigma \delta_i)_y$$

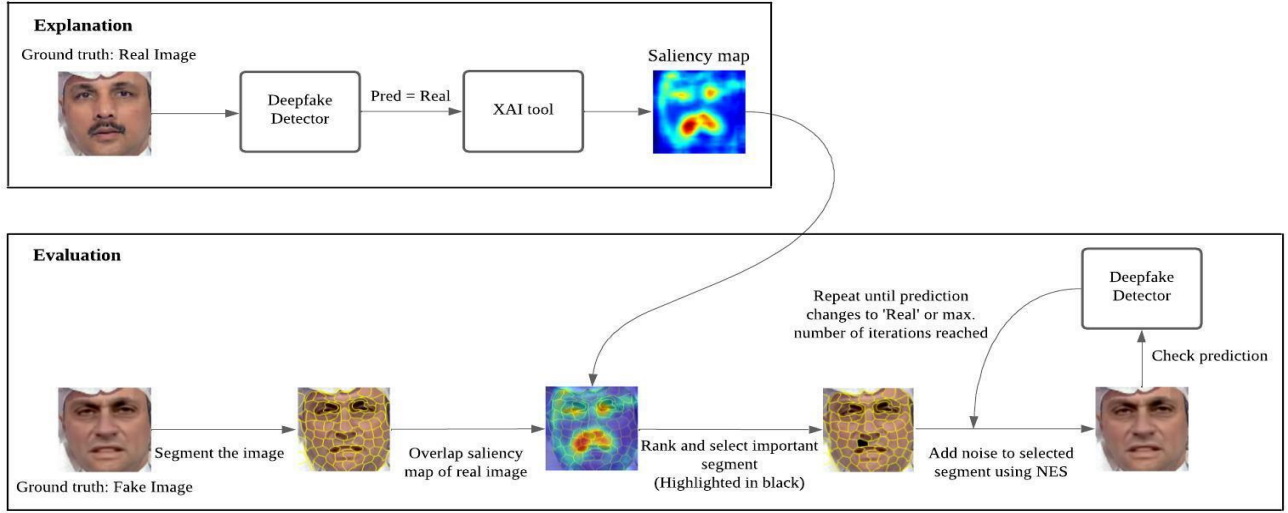

Fig. 4: Evaluation framework

Since our approach deals with adding noise to only a particular segment instead of the entire image, we modify the above equation as follows:

$$\nabla E[F(\theta)] \approx \frac{1}{\sigma n} \sum_{i=1}^{n} \delta_i[ind] F(\theta[ind] + \sigma \delta_i[ind])_y$$

After estimating the gradient, the input is iteratively moved in the direction of this gradient using gradient sign updates to increase the probability of target class:

$$x_i[ind] = x_{i-1}[ind] + \text{clip}_\epsilon(\alpha \cdot \text{sign}(\nabla F(x_{i-1})_y))$$

The implementation of adversarial fake sample generation is given in Algorithm 1.

## 5. EXPERIMENTAL SETUP

All the experiments were done using python 3.9.12 on a 64 GB RAM Ubuntu 22.04 OS machine with a NVIDIA GeForce RTX 3070 GPU.

### 5.1. Dataset and Models

FaceForensics++ [30] and Celeb-DF [31] datasets were used for our experiments. FaceForensics++ is a collection of fake videos created using four manipulation methods: Deepfake (DF), Face2Face (F2F), FaceSwap (FS) and NeuralTextures (NT). There are 1000 real videos and each manipulation method consists of 1000 videos. The videos are offered in different compression modes. We chose the raw format for our experiments. The dataset is divided into 720 videos for training and 140 videos each for validation and testing. For the 140 videos in the test set, we sampled 10 frames from each video. Thus, a total of 1400 original and fake images were used to carry out our proposed method. Celeb-DF consists of 590 real videos and 5639 fake videos. The test set consists of 518 videos in total. We sampled 5 frames from this larger test set. Kindly note that while Celeb-DF is open-sourced and publicly available, we cannot distribute any derived or manipulated data from the dataset according to their terms of use as listed in [32].

Two factors had an impact on our choice for the number of sampled frames from the datasets. The datasets have a frame rate of 30 fps(frames per second). Given that there are 140 and 518 videos in the test set of FaceForensics++ and Celeb-DF datasets respectively, working with every frame will increase computational and time complexity as highlighted by recent works [33] and [34]. To reduce computational complexity, [33] samples only 4 frames for training models while [34] uses 8 frames. The second reason is the fact that recent deepfake detectors rely on finding spatial inconsistency among frames [35]. Thus, an attacker would effectively need to modify the same spatial region across multiple frames to bypass detection. Given that the datasets have a frame rate of 30 fps with an average video length of 13 seconds [36], the sampling of 5-10 frames across the video should be sufficient to demonstrate that the entire video is vulnerable.

We implemented our approach on MesoNet [37] and XceptionNet. We have highlighted few issues and the reasons for our experimental choices below:

- XceptionNet accepts images of any size whereas MesoNet accepts images of size 256 x 256 x 3 only. While XceptionNet's publicly available model was originally trained on images of size 299 x 299 x 3, we have carried out our experiments for size 256 x 256 x 3 to maintain consistency between both models.
- Out of the four manipulation methods provided by FaceForensics++, we have tested out our evaluation proposal on DF, F2F and FS only. The accuracy of XceptionNet on 1400 real images for NT was just 20%. We believe that this could be due to the fact that NT manipulates only few frames in the source video and thus, 1400 frames may not be enough to get a good accuracy on NT.

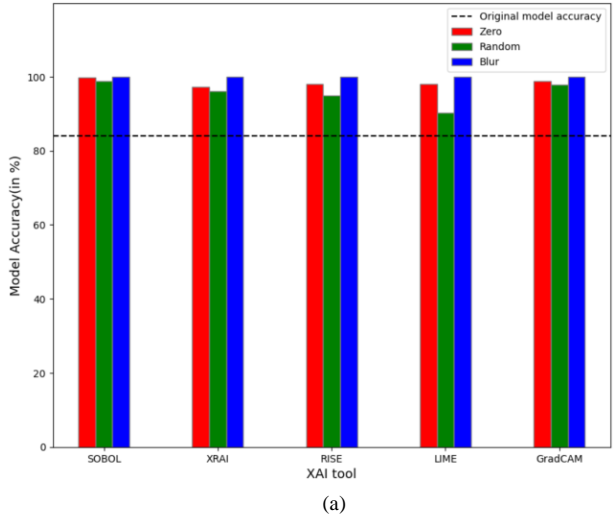 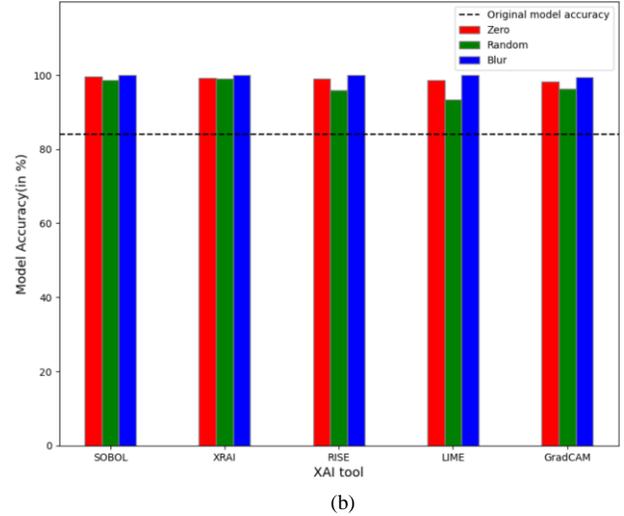

(a)      (b)

Fig. 5: Model accuracy results for removal XAI evaluation metrics on XceptionNet for FaceForensics++ DF (a) Deletion: Top 15% of salient pixels replaced with zero, random and blurred values (b) IROF: Top 15 salient segments replaced with zero, random and blurred values

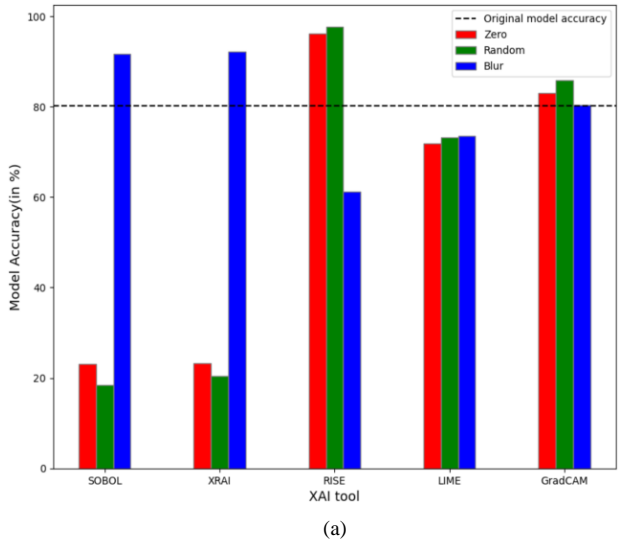 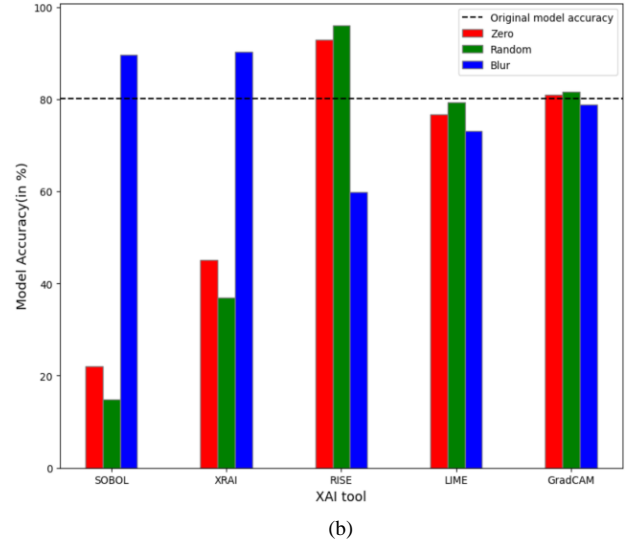

(a)      (b)

Fig. 6: Model accuracy results for removal XAI evaluation metrics on XceptionNet for Celeb-DF (a) Deletion: Top 15% of salient pixels replaced with zero, random and blurred values (b) IROF: Top 15 salient segments replaced with zero, random and blurred values

- The pre-trained models of MesoNet and XceptionNet were available in different versions. In the case of MesoNet, three models trained individually on DF, F2F and FS and one model trained on all manipulation methods was available. However, in the case of XceptionNet, only the model trained on all manipulation methods was available. The accuracy of 1400 real images on the MesoNet model that was trained on all manipulation methods was only 38.36%. In such a case, the explanations retrieved for real images cannot be trusted upon since the prediction probability itself will point more towards the fake class. Hence for MesoNet, we have used the three model versions individually trained on DF, F2F and FS as opposed to XceptionNet where the model trained on all manipulation methods was used.

### 5.2. Configuration of XAI tools

As each tool has various parameters, we would like to highlight the configuration used for each tool since different configurations can produce varying results.

1) SOBOL: For both models, the default values were used for each parameter.

2) XRAI: For both models, the default values were used for each parameter.

3) RISE: There is no default value for the number of masks in RISE. Since RISE is computation heavy, we limited the number of masks to 2000.

4) LIME: The default value of 1000 perturbations was used to create explanations for both models. LIME makes use of segmentation algorithms built on scikit-image to perturb different segments and ranks them based on their importance. We made use of the slic algorithm for LIME as well.

5) Grad-CAM: Any CNN layer can be utilized to view Grad-CAM's explanation, but in practice it is most preferred to use the last CNN layer of a model. Hence, 'conv2d 15' and 'conv4.pointwise' layers were used to visualize MesoNet and XceptionNet respectively.

*5.3. Configuration of NES*

1) Maximum iterations *itr*: The number of iterations was set as 50 for both models. This number was chosen to keep the computation time of the evaluation process as less as possible while at the same time providing sufficient amount of trials for NES to reach an adversarial solution.

2) Learning rate $\alpha$: The learning rate was set as $\frac{1}{255}$. The reason for this is to make sure that NES does not add too much noise at each step. If too much noise is added at each step, then there could be a possibility where the prediction probability of the adversarial fake image never comes close to 0 but rather moves in the opposite direction.

3) Maximum distortion $\epsilon$: The maximum amount of noise added to an image was capped at $\frac{16}{255}$ to maintain visual imperceptibility. It might be easier to spot the distortion done to pixels if this number is increased.

4) Search variance $\sigma$: The search variance was set as 0.001 following the implementation of [29].

5) Number of samples *n*: This was set as 20 for MesoNet and 80 for XceptionNet. The number of samples had to be increased for XceptionNet due to its complexity when compared to MesoNet.

## 6. RESULTS

In this section, we first demonstrate the limitations of the applicability of existing removal/insertion XAI evaluation methods on deepfake detectors. We then show the results of our implemented approach.

*6.1. Analysis of generic removal/insertion XAI evaluation methods*

We investigated the results of different removal/insertion XAI evaluation methods on XceptionNet for FaceForensics++ DF and Celeb-DF datasets. The chosen metrics were Deletion, IROF and IAUC.

In Deletion, top 15% of salient pixels highlighted by an XAI tool are replaced with either a zero value, uniform random value or blurred using neighbouring pixels. The model accuracy is expected to drop after replacement and the magnitude of drop determines the effectiveness of the tool. IROF is similar to Deletion, with the only difference being the replacement of 15 salient segments instead of 15% of salient pixels. In IAUC, top 15% of salient pixels are inserted on a completely blurred version of the original image. The model accuracy is expected to rise after the insertion and the magnitude of rise determines the effectiveness of the tool.

Figure 5(a) and 5(b) show the results of Deletion and IROF for XceptionNet on FaceForensics++ DF. It can be observed that none of the tools show a drop in model accuracy on computation of Deletion and IROF. Rather interestingly, the numbers are greater than the actual accuracy of 84% which is represented by the dashed line. This shows that replacement of pixels on fake images is not suitable to evaluate XAI tools since the detectors rely on face artifacts to perform detection. Replacement of pixels/segments results in distorting those face artifacts and can produce unexpected results as can be seen in this case.

Figure 6(a) and 6(b) show the results of removal metrics for XceptionNet on Celeb-DF. While the results of some tools fall below the actual accuracy of 80.11%, there are still some abnormal values which make the results difficult to compare. For instance, the results for RISE and GradCAM are well above 80.11% in most of the cases. This raises questions on the validity of these metrics since the model accuracy is expected to drop on removal of pixels/segments. Since the results of RISE and GradCAM do not show a drop in model accuracy, it also makes us question whether the drop in the results of other tools was genuine or random.

Table I shows the results of IAUC for XceptionNet on the two datasets. The second row shows the accuracy of XceptionNet on the blurred versions of the respective datasets. When the top 15% of salient pixels are inserted onto these blurred images, the accuracy is expected to increase. However, the results were not as expected. For FaceForensics++ DF, the results of all tools remained at 0% while for Celeb-DF, the results of SOBOL, XRAI and GradCAM showed a decrease in accuracy. The results are not meaningful enough to evaluate the tools since they deviate from the expected behaviour of an increase in accuracy. This also shows that blurring may not be an effective replacement method to evaluate XAI tools on deepfake images.

Overall, the results of our experiments show that generic removal/insertion XAI evaluation methods may or may not work well for specific image processing tasks. This warrants the need to research and develop new XAI evaluation methods specific to the task that the model has been trained for.

| Dataset | FaceForensics++ DF | Celeb-DF |
|---|---|---|
| **Accuracy** (Blurred images) | 0% | 74.82% |
| **SOBOL** | 0% | 20.58% |
| **XRAI** | 0% | 21.29% |
| **RISE** | 0% | 97.17% |
| **LIME** | 0% | 78.23% |
| **GradCAM** | 0% | 62.28% |

TABLE I: IAUC results on XceptionNet. The second row shows the accuracy of the model on the blurred version of the dataset. Rows 3-7 show the accuracy of the model on inserting the top 15% of salient pixels onto the blurred images. The values in rows 3-7 should be greater than the values in row 2.

*6.2. Analysis of our proposed approach*

Table II shows the results of our implementation on FaceForensics++ dataset. In the case of MesoNet, we can see that XRAI is the most faithful tool for DF whereas LIME and RISE perform well on F2F and FS respectively. Note that since we have used three different MesoNet models trained

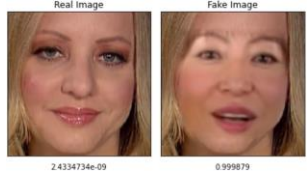

(a) The prediction of MesoNet on a real-fake(DF) image pair (A real image has a probability close to 0 while a fake image has a probability close to 1)

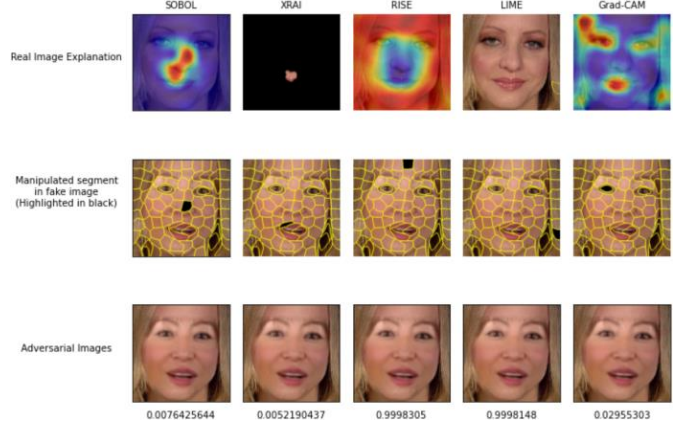

(b) The prediction of MesoNet on the adversarial DF images generated using the results of different XAI tools

Fig. 7: An example of the evaluation on MesoNet

individually on DF, F2F and FS, the faithfulness of tools will not be similar for all of them. For instance, XRAI which performed well on the MesoNet model trained on DF, had the least drop in model accuracy for adversarial F2F images. However, this is not the case with XceptionNet. Since the evaluation is carried out on a single model trained on all manipulation methods, the faithfulness of tools should be the same regardless of whichever dataset is used. Our experiment results prove the same. GradCAM showed the best results for DF, F2F and FS whereas LIME had the least drop in model accuracy for adversarial images of all datasets. Figure 7 shows an example of the evaluation process for a real-fake image pair on MesoNet.

Table III shows the results of our approach for XceptionNet on Celeb-DF. We could not use Celeb-DF with MesoNet as a pre-trained model was not available publicly. XRAI showed the best results while the adversarial images created using RISE showed the highest adversarial accuracy.

|  | MesoNet | | | XceptionNet | | |
| --- | --- | --- | --- | --- | --- | --- |
|  | DF | F2F | FS | DF | F2F | FS |
| Original Accuracy | 95.5% | 53.21% | 83.49% | 84% | 94.42% | 98.85% |
| SOBOL | 5.64% | 29.35% | 74.28% | 53.14% | 65.57% | 88% |
| XRAI | 5.5% | 38.28% | 72.07% | 57.07% | 75.64% | 89.64% |
| RISE | 88.07% | 29.85% | 20.71% | 58.5% | 72.28% | 88.42% |
| LIME | 71.71% | 19.92% | 45.85% | 61.57% | 80% | 92.14% |
| GradCAM | 54.07% | 44.28% | 46.78% | 32.28% | 42.78% | 83% |

TABLE II: Implementation of our approach on FaceForensics++ dataset. The original accuracy of the models on actual fake images is reported in the third row. The accuracy of the models on adversarial fake images created using the respective tools is reported in rows 4-8. A faithful tool should reduce the accuracy of a model on adversarial fake samples.

| Original Accuracy | 80.11% |
| --- | --- |
| SOBOL | 27.29% |
| XRAI | 19.64% |
| RISE | 28.58% |
| LIME | 25.17% |
| GradCAM | 21.94% |

TABLE III: Implementation of our approach on XceptionNet for Celeb-DF dataset

We have also outlined a comparison of the evaluation results on XceptionNet between images of size 299 x 299 x 3 and 256 x 256 x 3 in Table IV. We can observe that the ability to generate adversarial images reduces when the image resolution is increased.

Do note that we are manipulating only one segment in each image. Since we are dividing a total of 65536 pixels into 100 segments, roughly around 1% of pixels gets manipulated to generate an adversarial image. The accuracy of the models on adversarial fake samples may decrease even further if more segments are manipulated. Our main objective was to find out the most faithful tool by keeping the amount of distortion on an image as less and imperceptible as possible.

Table V and VI show the average time taken for computation of explanation and evaluation by different tools respectively. As far as time complexity of explanations is concerned, we can observe that Grad-CAM seems to provide quick responses in general. This can be attributed to the fact that it involves relatively simple computations when compared to the other tools. While the time taken by model-specific tools like Grad-CAM and XRAI may depend on the model's structure, the time taken by model-agnostic tools on different models should be similar for a given set of parameters since they do not depend upon the model's structure. However, a significant increase in the amount of explanation time for SOBOL and RISE can be observed on XceptionNet. We had to reduce the batch size of images while applying SOBOL and RISE on XceptionNet to make them fit our GPU. Thus, based on our observation, lowering of batch size of images may lead to an increased computation time for perturbation based tools.

| Image Size | 256 x 256x3 | 299 x 299x3 |
|---|---|---|
| Original Accuracy | 84% | 94.2% |
| SOBOL | 53.14% | 71.07% |
| XRAI | 57.07% | 76.57% |
| RISE | 58.5% | 73.71% |
| LIME | 61.57% | 75.28% |
| GradCAM | 32.28% | 64.5% |

TABLE IV: Accuracy of adversarial DF images created using respective tools on XceptionNet

| | MesoNet | XceptionNet |
|---|---|---|
| SOBOL | 1.17 | 5.07 |
| XRAI | 3.24 | 32.6 |
| RISE | 12 | 23.54 |
| LIME | 5.73 | 5.56 |
| GradCAM | 4.94 | 2.52 |

TABLE V: Average time taken (in seconds) by respective tools for explaining one prediction by the models

| | MesoNet | | | XceptionNet | | |
|---|---|---|---|---|---|---|
| | DF | F2F | FS | DF | F2F | FS |
| SOBOL | 5.64% | 29.35% | 74.28% | 53.14% | 65.57% | 88% |
| XRAI | 5.5% | 38.28% | 72.07% | 57.07% | 75.64% | 89.64% |
| RISE | 88.07% | 29.85% | 20.71% | 58.5% | 72.28% | 88.42% |
| LIME | 71.71% | 19.92% | 45.85% | 61.57% | 80% | 92.14% |
| GradCAM | 54.07% | 44.28% | 46.78% | 32.28% | 42.78% | 83% |

TABLE VI: Average time taken (in seconds) for generating one adversarial fake image using the segment highlighted by the respective tools

## 7. Conclusion

Due to the rapid spread of deepfake content across the internet, numerous deep learning models are being proposed to distinguish between real and fake content. While a model may claim a high detection accuracy, it is also important to gain the trust of the person deploying it. XAI tools have played a tremendous role in the last few years by helping humans understand the working of a model. However, blindly using an XAI tool's result to trust/mistrust the model is also not recommended given that different XAI tools work using different strategies. It is important to ensure that an XAI tool remains faithful to a model. In this paper, we demonstrated the limitations of existing removal/insertion XAI evaluation methods on deepfake detectors. We also presented a novel approach to evaluate the faithfulness of an XAI tool on a deepfake detection model. We proposed to evaluate tools based on their ability to generate adversarial fake images using the explanation of corresponding real images. We believe that this approach will aid researchers and developers to deploy the right tools on their models.

One limitation of our work is that it requires the presence of a corresponding real video for a fake video, meaning that fake videos that do not have a real video counterpart cannot be used with our approach. This drawback hindered us from exploring other deepfake datasets like UADFV and DFDC. Another major limitation of our approach is that it may not work on deepfake detectors that are robust to adversarial attacks. Adversarially robust detectors cleverly neglect the noise added to images and continue to give the same prediction regardless of the distortion. Developing evaluation methods that work on adversarially robust detectors can be explored as future work.